\documentclass{article} 
\usepackage{iclr2027_conference,times}

\usepackage{amsmath,amsfonts,bm}

\def\eqref#1{equation~\ref{#1}}

\def\1{\bm{1}}

\DeclareMathAlphabet{\mathsfit}{\encodingdefault}{\sfdefault}{m}{sl}
\SetMathAlphabet{\mathsfit}{bold}{\encodingdefault}{\sfdefault}{bx}{n}

\usepackage{hyperref}
\usepackage{url}
\usepackage{graphicx}
\usepackage{amsmath}
\usepackage[ruled,vlined,linesnumbered]{algorithm2e}
\usepackage{multirow}
\usepackage{subcaption}
\title{MAPLE: MoE Adaptive Plug-and-play Layer-wise Expert allocation}

\author{
Lie Li\\
University of Bristol\\
\texttt{fx260102@bristol.ac.uk}\\
\And
Wen Li\\
University of Bristol\\
\texttt{wen.li@bristol.ac.uk}\\
\And
Junxiao Shen\\
University of Bristol\\
junxiao.shen@bristol.ac.uk\\
\And
Guosheng Hu\\
University of Bristol\\
g.hu@bristol.ac.uk\\
}

\iclrfinalcopy 
\begin{document}

\maketitle

\begin{abstract}
Sparsely-activated Mixture-of-Experts (MoE) Transformers universally fix the same number of routed experts across all layers, a convention that ignores the well-documented heterogeneity in layer-wise redundancy. We demonstrate that this uniformity is systematically suboptimal and propose \textbf{MAPLE}, a plug-and-play framework that reallocates the routed-expert budget heterogeneously across layers of any pretrained MoE LLM, without modifying weights or requiring retraining. Our core contribution is a closed-form sensitivity-guided allocation: we probe each layer's response to variation in expert count, quantify sensitivity using three measures, and derive an analytically optimal budget assignment that directs capacity towards sensitive layers and absorbs reductions in redundant layers. This closed-form solution is further refined by a sensitivity-constrained genetic search that uses layer-wise sensitivity as a before guide for exploration, yielding faster convergence and superior allocation quality. On four MoE models spanning different scales and architectures, \textbf{MAPLE} outperforms uniform and pruning-based baselines under a 75\% routed-expert budget. Notably, on DeepSeek-MoE-16B, \textbf{MAPLE} uses only 75\% of the experts yet surpasses the original 100\% expert-uniform baseline on ARC-E, ARC-C, and BoolQ, improving accuracy from 65.09 to 71.40, 48.49 to 51.50, and 80.03 to 82.38, respectively. These accuracy gains translate into measured deployment efficiency: implementing \textbf{MAPLE} in SGLang reduces single-GPU end-to-end serving latency by 32.2\% and improves throughput by 47.4\%. These results show that well-designed heterogeneous allocation can be more effective than simply activating more experts, establishing it as a principled and practical axis for improving MoE efficiency.
\end{abstract}
\begin{figure}[h]
    \centering
    \includegraphics[width=\linewidth]{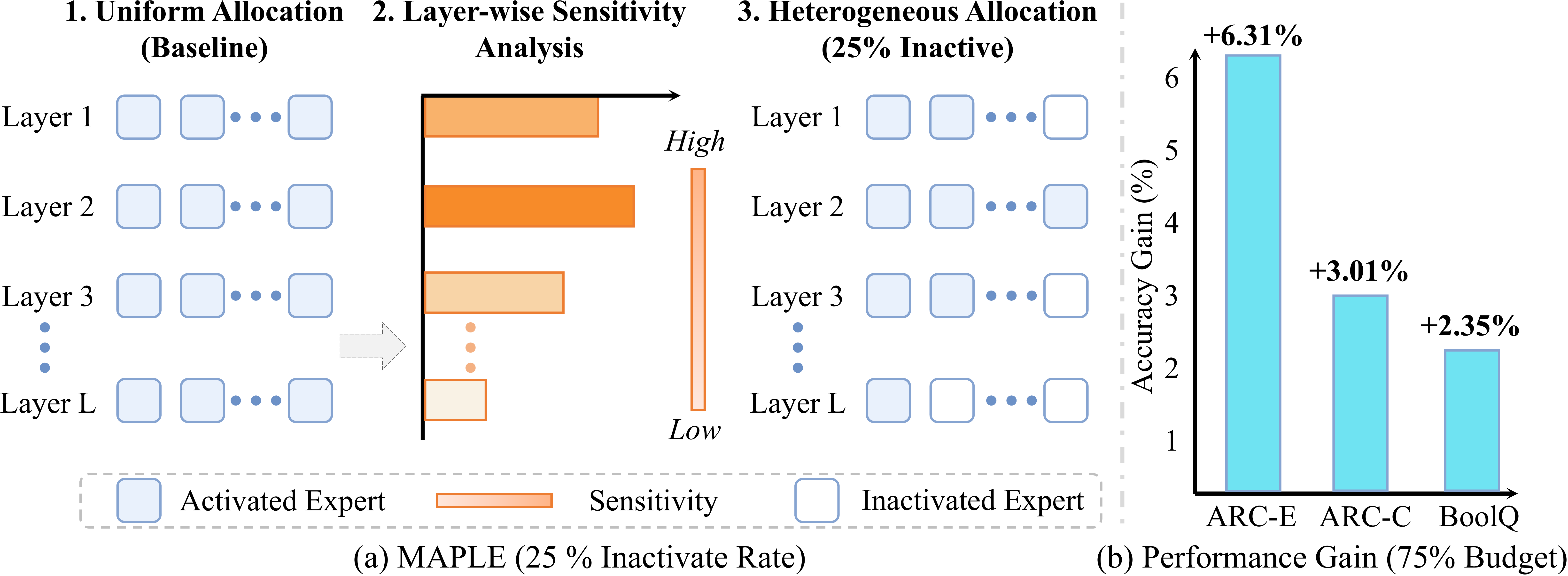}
  \caption{
Main idea and key result of \textbf{MAPLE}. 
\textbf{(a)} Overview of  \textbf{MAPLE}: uniform expert allocation is replaced by sensitivity-guided heterogeneous allocation under the same budget. 
\textbf{(b)} Less compute, better performance: with only 75\% routed experts,  \textbf{MAPLE} surpasses the full-expert uniform baseline on 
ARC-Easy (ARC-E), ARC-Challenge (ARC-C), and BoolQ reasoning benchmarks.
}
    \label{fig:maple_main}
\end{figure}

\section{Introduction}

Plug-and-play adaptability, first formalised in signal processing \citep{venkatakrishnan2013pnp}, describes the ability to insert a module into an existing system and immediately reap benefits without redesigning it from scratch. Deep learning has demonstrated this principle through several landmark interventions: Batch Normalization \citep{ioffe2015batchnorm} stabilised training across diverse architectures; Residual connections \citep{he2016resnet} unlocked networks an order of magnitude deeper; Dropout \citep{srivastava2014dropout} became a widely applicable regulariser requiring minimal modification. In each case, the intervention was lightweight yet delivered outsized returns.

As neural networks developed, the plug-and-play philosophy also became more targeted. Rather than applying the same intervention uniformly across an entire model, researchers began introducing lightweight modules that selectively emphasised the most informative components. In convolutional neural networks, Squeeze-and-Excitation blocks~\citep{hu2018senet} and CBAM~\citep{woo2018cbam} exemplified this shift by adaptively recalibrating feature responses through simple add-on modules. The broader lesson was clear: when internal components contribute unevenly, targeted modular interventions can be more effective than uniform ones.

As Transformer-based language models continued to scale~\citep{devlin2019bert, raffel2020t5, brown2020gpt3, chowdhery2022palm, touvron2023llama, touvron2023llama2}, full retraining became increasingly expensive, making plug-and-play adaptation more attractive. Adapter layers~\citep{houlsby2019adapter}, LoRA~\citep{hu2022lora}, Prefix tuning~\citep{li2021prefix}, prompt tuning~\citep{lester2021prompt}, and BitFit~\citep{zaken2022bitfit} offered efficient alternatives by updating only a small subset of parameters. Yet despite evidence that different model components contribute unequally~\citep{bahdanau2015attention, vaswani2017attention, michel2019sixteen, voita2019analyzing}, these methods still largely treat adaptation uniformly rather than tailoring it to the parts of the model that matter most.

This untapped dimension becomes especially important in Mixture-of-Experts (MoE) LLMs~\citep{jacobs1991moe, shazeer2017outrageously}, where model capacity is exposed explicitly through sparse expert activation. By replacing dense feed-forward layers with sparsely activated expert pools, MoE models decouple parameter capacity from per-token compute~\citep{fedus2022switch, lepikhin2021gshard, du2022glam}. With continued improvements in stability and routing~\citep{zoph2022stmoe, zhou2022mixture}, MoE architectures have also been adopted in a number of recent high-performing models, including OLMoE~\citep{muennighoff2024olmoe}, DeepSeek-MoE~\citep{dai2024deepseekmoe}, and Moonlight~\citep{liu2025moonlight}. Yet despite this flexibility, current pretrained MoE LLMs still typically use a fixed per-layer number of active experts, even though recent work suggests that expert importance and redundancy vary substantially across layers~\citep{guo2023sensitivity, lu2024not, zhang2025mone}.

Existing evidence also points to substantial layer-wise heterogeneity. In parameter-efficient fine-tuning, higher layers have been shown to benefit from more LoRA experts~\citep{gao2024mola}, suggesting that adaptation demands are not uniform across depth. More importantly, recent studies on expert pruning and layer compression indicate that redundancy in MoE models varies substantially across layers~\citep{lu2024not, liu2024efficient, easyep2025, cao2024condense}. Taken together, these findings suggest that routed expert capacity should not be allocated uniformly, but instead determined in a layer-aware manner.

However, implementing such layer-aware allocation through existing approaches remains costly. Unstructured methods~\citep{han2015learning, frantar2023sparsegpt, sun2024wanda} and MoE-specific variants~\citep{xie2024moepruner} often rely on calibration data, weight reconstruction, or additional recovery procedures to maintain performance. Structured methods that permanently remove experts, whether guided by activation statistics~\citep{lu2024not}, low-rank decomposition~\citep{yang2024moei2}, condensation~\citep{cao2024condense}, Shapley-value estimation~\citep{shapley2024moe}, or domain-specific calibration~\citep{easyep2025}, similarly introduce extra data, retraining, or task-specific overhead. Expert merging~\citep{li2024mcsmoe} adds further computational complexity. In addition, representative MoE pruning methods such as MoNE~\citep{zhang2025mone} and EEP~\citep{liu2024efficient} also operate by identifying and removing redundant experts or expert parameters. Since these methods alter the original expert structure rather than only reallocating the active expert budget at inference time, they are less amenable to plug-and-play deployment on pretrained MoE models.

In this work, we propose \textbf{MAPLE} (\textbf{M}oE \textbf{A}daptive \textbf{P}lug-and-play \textbf{L}ayer-wise \textbf{E}xpert allocation), a plug-and-play framework for automatically determining heterogeneous per-layer expert allocations in pretrained MoE Transformers without modifying weights or retraining. \textbf{MAPLE} formulates expert reallocation as a constrained optimisation problem: it first performs a lightweight layer-wise sensitivity scan, then derives a closed-form heterogeneous allocation under a global expert budget, and subsequently applies a sensitivity-guided genetic refinement, all in a fully post-hoc manner without modifying model architectures. Figure~\ref{fig:maple_main} provides an overview of \textbf{MAPLE}, showing that sensitivity-guided heterogeneous expert allocation improves performance under a fixed budget and can even surpass the full-expert uniform baseline using only 75\% experts.

Our main contributions are as follows:
\begin{itemize}
    \item We identify heterogeneous per-layer expert allocation as an underexplored yet effective axis for improving pretrained MoE LLMs. Unlike existing work \citep{frantar2023sparsegpt,sun2024wanda,zhang2024RIA,liu2024efficient,zhang2025mone}, we  propose \textbf{MAPLE}, a fully plug-and-play framework that requires no weight updates, retraining.
    \item We introduce a sensitivity-guided optimisation strategy for expert reallocation, in which a lightweight layer-wise sensitivity scan enables a closed-form allocation under a global expert budget, with genetic search  for subsequent refinement.
    \item Under the same expert budget, \textbf{MAPLE} improves overall performance over the uniform baseline and delivers superior results compared with pruning-based and expert-reallocation methods across five reasoning benchmarks. In particular, even with 25\% fewer routed experts, \textbf{MAPLE} is able to surpass the original full-expert MoE on multiple tasks, highlighting the value of sensitivity-guided heterogeneous expert allocation.
\end{itemize}

\section{Related Work}

\noindent\textbf{{Layer-wise Sensitivity Analysis.}}
Understanding layer-wise response to capacity reduction is fundamental to model compression. Prior work has shown that network redundancy is substantial but highly non-uniform across layers~\cite{han2015learning, frankle2019lottery, guo2023sensitivity}. Sensitivity-based analyses therefore motivate non-uniform compression, where capacity is allocated according to functional importance rather than reduced uniformly~\cite{engelbrecht2001pruning, ma2023llmpruner}. 
Evidence from structured pruning and attention-head analysis further supports this view, showing that layers and sub-components contribute unequally to performance~\cite{michel2019sixteen, voita2019analyzing}.

\noindent\textbf{{Mixture-of-Experts Models and Expert Compression.}}
MoE models improve parameter efficiency through sparse expert activation~\citep{shazeer2017outrageously, fedus2022switch}, as shown by recent architectures such as Mixtral~\citep{jiang2024mixtral} and DeepSeekMoE~\citep{dai2024deepseekmoe}. To reduce deployment cost, early work adapted dense-model pruning methods to MoE experts, including SparseGPT~\citep{frantar2023sparsegpt}, Wanda~\citep{sun2024wanda}, and router-aware pruning in MoE-Pruner~\citep{xie2024moepruner}. Because unstructured sparsity often brings limited practical acceleration, later studies shifted to structured expert-level compression. These include activation-frequency-based pruning in Not All Experts are Equal~\citep{lu2024not}, evolutionary expert search in EEP~\citep{liu2024efficient}, and expert merging in MC-SMoE~\citep{li2024mcsmoe}. More recent methods further introduce lightweight replacements or calibration-guided compression, such as MoNE~\citep{zhang2025mone}, ConDense-MoE~\citep{cao2024condense} and domain-specific expert pruning approaches~\citep{easyep2025}. However, these methods generally modify model weights or structure, rely on calibration data, and often overlook heterogeneous layer-wise redundancy. LExI\citep{chittyvenkata2025lexilayeradaptiveactiveexperts} is more closely related to MAPLE: it also preserves the pretrained model and assigns heterogeneous active-expert counts across layers, but derives the allocation in a data-free manner from model-weight statistics. Our method instead reallocates the active expert budget across layers based on sensitivity, without altering model weights.

\noindent\textbf{{Budget-Constrained Architecture Search.}}
From an optimisation perspective, allocating a fixed budget non-uniformly across layers is a discrete combinatorial problem whose search space grows rapidly with network depth, making exact optimisation impractical for large models~\citep{elsken2019nas}. Prior work has therefore relied on approximate search strategies. In CNNs, AMC~\citep{he2018amc} used reinforcement learning to assign layer-wise compression ratios under a global FLOPs budget and showed clear gains over uniform policies. NetAdapt addressed the same problem with iterative greedy adaptation, trading global optimality for search efficiency~\citep{yang2018netadapt}. NAS later generalised this idea to broader architecture search, where evolutionary methods proved effective in discrete design spaces~\citep{elsken2019nas, real2019regularized, tan2019mnasnet}. Similar observations have also emerged in Transformers: MoLA showed that heterogeneous LoRA expert assignment outperforms uniform allocation under the same parameter budget~\citep{gao2024mola}. Our method follows this line of work, but avoids expensive search in the allocation step by deriving a closed-form layer-wise expert allocation from sensitivity estimates.
\section{Method}
We present \textbf{MAPLE}, a plug-and-play framework for sensitivity-guided heterogeneous expert allocation in budget-constrained MoE Transformers. Layers that are more sensitive to expert count should deviate less from their preferred allocation. MAPLE proceeds in three steps: it estimates layer-wise sensitivity and identifies each layer's single-layer optimum (Section~\ref{sec:sensitivity}); derives a closed-form allocation under the global budget (Section~\ref{sec:allocation}); and refines the discrete solution with a lightweight genetic search (Section~\ref{sec:refinement}).

\subsection{Layer-wise Sensitivity Modelling}
\label{sec:sensitivity}

MoE Transformers typically use a uniform expert count across layers, although layers play distinct roles and exhibit varying redundancy~\citep{guo2023sensitivity,lu2024not,zhang2025mone}. Because routing is performed independently at each layer~\citep{qiu2024rmoe}, we probe layers individually. Specifically, while fixing all other layers at the baseline expert count $k_0$, we vary the expert count of layer $l$ over $\mathcal{E}=\{1,\dots,k_0\}$, where $e\in\mathcal{E}$ denotes the number of routed experts assigned to layer $l$, and record the validation accuracy $A_l(e)$. The single-layer optimum is $k_l^* = \min\{e \mid A_l(e) = \max_{e' \in \mathcal{E}} A_l(e')\}$.

The optimum $k_l^*$ alone does not indicate how strongly layer $l$ responds to expert-count variation. Prior work in pruning, quantisation, and adaptive fine-tuning similarly reports layer-dependent sensitivity~\citep{lecun1990optimal,molchanov2019importance,dong2019hawq,dong2020hawqv2,DBLP:journals/corr/abs-2004-09602,zhang2023adalora}. We therefore consider three lightweight sensitivity measures: $s_l^{\mathrm{acc}} = \max_{e \in \mathcal{E}} A_l(e) - \min_{e \in \mathcal{E}} A_l(e)$, $s_l^{w} = \frac{1}{|W_l|}\sum_{W \in W_l}\mathrm{Var}(W)$, and $s_l^{\mathrm{out}} = \frac{1}{|\mathcal{E}|}\sum_{e \in \mathcal{E}}\mathrm{MSE}\bigl(H_l(e), H_l(k_0)\bigr)$. Here, $W_l$ is the set of parameter tensors in layer $l$, and $H_l(e)$ is its output when using $e$ experts. These metrics capture performance-, parameter-, and representation-level sensitivity, respectively. We use accuracy range by default and evaluate the alternatives in the ablation study; $s_l$ denotes the selected metric.

\subsection{Sensitivity-guided Closed-form Expert Allocation}
\label{sec:allocation}

Given the preferred expert counts $\{k_l^*\}_{l=1}^L$, sensitivities $\{s_l\}_{l=1}^L$, and global budget $K$, we seek an allocation $\mathbf{k}=[k_1,\dots,k_L]$ that stays close to each layer's optimum. More sensitive layers should deviate less, so we solve $\min_{\mathbf{k}} \; \mathcal{J}(\mathbf{k}) = \sum_{l=1}^{L} w_l \bigl(k_l-k_l^*\bigr)^2 \quad \text{s.t.} \quad \sum_{l=1}^{L} k_l = K$, where $w_l=s_l>0$ in our implementation.

Introducing a Lagrange multiplier $\lambda$ gives $\mathcal{L}(\mathbf{k},\lambda) = \mathcal{J}(\mathbf{k}) + \lambda\!\left(\sum_{l=1}^{L} k_l-K\right)$. Setting $\partial\mathcal{L}/\partial k_l=0$ yields $k_l^{\mathrm{cont}} = k_l^* - \frac{\lambda}{2w_l}$, with $\lambda = \frac{2\!\left(\sum_{l=1}^{L} k_l^*-K\right)}{\sum_{l=1}^{L} 1/w_l}$.

Let $\Delta=K-\sum_l k_l^*$ and $\Gamma=\sum_l k_l^*-K$. Any surplus or deficit relative to $K$ is assigned preferentially to low-sensitivity layers:
\begin{equation}
    k_l^{\mathrm{cont}} =
    \begin{cases}
        k_l^* + \Delta\cdot \dfrac{1/s_l}{\sum_{j=1}^{L}1/s_j}, & \text{if } \sum_l k_l^* < K, \\[2mm]
        k_l^* - \Gamma\cdot \dfrac{1/s_l}{\sum_{j=1}^{L}1/s_j}, & \text{if } \sum_l k_l^* > K.
    \end{cases}
    \label{eq:budget_case}
\end{equation}

We clip $\mathbf{k}^{\mathrm{cont}}$ to the feasible range $[k_{\min},k_{\max}]$ and apply the largest-remainder method to obtain an integer allocation $\mathbf{k}^{\mathrm{sens}}$ satisfying the per-layer bounds and global budget.

\subsection{Sensitivity-constrained Genetic Refinement}
\label{sec:refinement}

The closed-form solution $\mathbf{k}^{\mathrm{sens}}$ provides a principled initialization but may retain rounding artefacts and residual cross-layer interactions. We therefore apply a lightweight genetic algorithm with validation accuracy $A(\mathbf{k})$ as fitness. Candidate allocations are generated by sensitivity-controlled Gaussian mutation: $\tilde{k}_l = k_l^{\mathrm{sens}} + \epsilon_l, \quad \epsilon_l \sim \mathcal{N}(0,\sigma_l^2), \quad \sigma_l^2 \propto \frac{1}{s_l+\varepsilon}$, so high-sensitivity layers receive smaller perturbations. At each generation, we retain the top two candidates as elites and generate the remaining population using this same perturbation, without crossover. The best candidate gives the final allocation $\mathbf{k}^{\mathrm{best}}$.

\subsection{Deployment Compatibility}
\label{sec:deployment}

Because \textbf{MAPLE} only changes, per MoE layer, how many router-ranked experts are activated at inference time, without modifying the router, expert weights, or model architecture, it can be integrated into any MoE serving engine that exposes a per-layer top-$k$ routing parameter. We implement \textbf{MAPLE} in SGLang by introducing a dedicated execution path for layers assigned $k=1$: after the router computes per-expert scores, only the highest-scoring (argmax) expert is evaluated, bypassing the multi-expert dispatch, weighted combination, and reduction logic required when $k>1$. Section~\ref{sec:experiments} reports the resulting end-to-end latency and throughput gains measured in this SGLang implementation. The same mechanism applies to vLLM's fused MoE kernel, which similarly exposes top-$k$ as a per-layer configuration parameter, so we expect comparable integration to be possible without further modification to \textbf{MAPLE} itself.

\section{Experiments}
\label{sec:experiments}
\subsection{Experimental Settings}
\paragraph{Model architectures.}
We evaluate \textbf{MAPLE} on four representative MoE models: DeepSeek-MoE-16B\citep{dai2024deepseekmoe}, DeepSeek-V2-Lite\citep{deepseekv2}, Moonlight-16B-A3B\citep{liu2025moonlight}, and OLMoE-1B-7B\citep{muennighoff2024olmoe}. To assess expert reallocation under constrained resources, we retain 75\% of the routed experts in each model while preserving the number of shared experts when present. As described in Section~\ref{sec:sensitivity}, the sensitivity scan and allocation search operate only over each model's MoE layers, and all reported budgets (75\% routed budget / full routed+shared budget, computed over these layers only) are: DeepSeek-MoE-16B 108/162+54; DeepSeek-V2-Lite and Moonlight-16B-A3B 104/156+52; and OLMoE-1B-7B, which has no shared experts, 96/128+0.

\paragraph{Evaluation benchmarks.}
We evaluate performance on seven benchmarks spanning both short-form MCQA and long-form generative reasoning. The MCQA benchmarks include RTE for textual entailment, BoolQ for yes/no question answering, PIQA for physical commonsense reasoning, and ARC-E/ARC-C for science question answering. We further include long-form generative benchmarks: GSM8K for mathematical reasoning with 0-shot CoT and BBH for challenging multi-step reasoning with 3-shot CoT.

\paragraph{Implementation and deployment details.}
All experiments use NVIDIA RTX A6000 GPUs with 48 GB memory and a 75\% routed-expert budget, with accuracy range as the default sensitivity metric. \textbf{MAPLE} computes a sensitivity-guided allocation and refines it using 10 candidates over 10 generations, retaining the top two elites per generation while enforcing the global budget and per-layer bounds. For end-to-end serving, we implement \textbf{MAPLE} in SGLang~0.4.6 and evaluate single-GPU, two-GPU pipeline-parallel (PP), and two-GPU expert-parallel (EP) configurations. Latency and throughput are averaged over five runs.

\subsection{Experimental Results}										

\paragraph{Overall performance.}

According to Table\ref{tab:fixed_budget_results}, \textbf{MAPLE} improves average accuracy over the full-budget baseline by 2.63, 6.44, and 4.13 points on DeepSeek-MoE-16B, DeepSeek-V2-Lite, and OLMoE-1B-7B, respectively, while using only 75\% of the routed experts; on Moonlight-16B-A3B, it is 0.62 points below baseline but still the best-performing method at this budget. Compared with the strongest baseline in each column, \textbf{MAPLE} improves average accuracy by 3.52, 5.79, 1.14, and 3.61 points across the four models, achieving the best average accuracy on every model evaluated. These results show that uniform expert allocation is often suboptimal, and that sensitivity-guided heterogeneous allocation can match or exceed it with fewer active experts.

\begin{table}[ht]
\caption{Main results under the 75\% routed-expert budget on four MoE architectures and five reasoning benchmarks. Baseline denotes the original full-budget uniform routed-expert configuration. Bold indicates the best result in each model--task block.}
\label{tab:fixed_budget_results}
\centering
\scriptsize
\setlength{\tabcolsep}{2.2pt}
\renewcommand{\arraystretch}{1.03}
\begin{tabular}{@{}lcccccc|cccccc@{}}
\hline
& \multicolumn{6}{c}{\textbf{DeepSeek-MoE-16B}} & \multicolumn{6}{c}{\textbf{DeepSeek-V2-Lite}} \\
\hline
\textbf{Method} & \textbf{ARC-C} & \textbf{ARC-E} & \textbf{BoolQ} & \textbf{PIQA} & \textbf{RTE} & \textbf{Avg.} & \textbf{ARC-C} & \textbf{ARC-E} & \textbf{BoolQ} & \textbf{PIQA} & \textbf{RTE} & \textbf{Avg.} \\
\hline
Baseline  & 48.49 & 65.09 & 80.03 & 80.09 & 84.84 & 71.71 & 44.82 & 64.74 & 67.40 & 79.82 & 53.43 & 62.04 \\
SparseGPT & 46.15 & 62.11 & 79.69 & 79.33 & 81.22 & 69.70 & 44.48 & 64.91 & 67.34 & 78.98 & 57.76 & 62.69 \\
Wanda     & 45.81 & 62.98 & 79.72 & 79.27 & 82.67 & 70.09 & 44.15 & 64.91 & 67.31 & 78.67 & 54.51 & 61.91 \\
RIA       & 46.48 & 63.16 & 79.72 & 79.49 & 80.87 & 69.94 & 45.15 & 64.04 & 65.08 & 78.99 & 49.82 & 60.62 \\
EPP       & 47.49 & 66.14 & 79.60 & 77.86 & 83.03 & 70.82 & 42.47 & 62.46 & 67.74 & 78.40 & 49.82 & 60.18 \\
MoNE      & 47.15 & 64.04 & 76.52 & 76.44 & 81.46 & 69.12 & 42.81 & 62.11 & 65.86 & 78.30 & 53.42 & 60.50 \\
LExI      & 47.49 & 67.19 & 79.85 & 78.35 & 74.37 & 69.45 & 47.16 & 64.21 & 68.78 & 76.82 & 54.15 & 62.54 \\
MAPLE     & \textbf{51.50} & \textbf{71.40} & \textbf{82.38} & \textbf{80.85} & \textbf{85.55} & \textbf{74.34} & \textbf{50.84} & \textbf{68.60} & \textbf{73.43} & \textbf{80.20} & \textbf{69.31} & \textbf{68.48} \\
\hline
& \multicolumn{6}{c}{\textbf{Moonlight-16B-A3B}} & \multicolumn{6}{c}{\textbf{OLMoE-1B-7B}} \\
\hline
\textbf{Method} & \textbf{ARC-C} & \textbf{ARC-E} & \textbf{BoolQ} & \textbf{PIQA} & \textbf{RTE} & \textbf{Avg.} & \textbf{ARC-C} & \textbf{ARC-E} & \textbf{BoolQ} & \textbf{PIQA} & \textbf{RTE} & \textbf{Avg.} \\
\hline
Baseline  & 77.26 & 88.60 & 76.54 & 79.71 & 69.68 & 78.36 & 41.47 & 54.91 & 65.96 & 78.35 & 77.98 & 63.73 \\
SparseGPT & 75.25 & 85.79 & 75.14 & 78.51 & 65.34 & 76.01 & 43.47 & 55.09 & 67.06 & 77.37 & 77.61 & 64.12 \\
Wanda     & 73.58 & 85.61 & 75.08 & 78.18 & 64.62 & 75.41 & 43.48 & 54.61 & 66.88 & 77.20 & 77.26 & 63.89 \\
RIA       & 76.25 & 84.25 & 74.20 & 78.45 & 68.31 & 76.29 & 43.14 & 55.96 & 64.74 & 78.34 & 79.06 & 64.25 \\
EPP       & 76.60 & 85.82 & 74.50 & 79.65 & 66.43 & 76.60 & 43.83 & 57.19 & 66.45 & 75.19 & 76.17 & 63.77 \\
MoNE      & 74.09 & 85.80 & \textbf{77.37} & 77.92 & 65.53 & 76.14 & 33.78 & 46.15 & 62.41 & 76.18 & 77.62 & 59.23 \\
LExI      & 67.89 & 81.58 & 70.61 & 78.67 & \textbf{71.12} & 73.98 & 30.43 & 38.95 & 63.73 & 76.82 & 53.07 & 52.60 \\
MAPLE     & \textbf{77.59} & \textbf{86.26} & 74.65 & \textbf{80.16} & 70.04 & \textbf{77.74} & \textbf{44.81} & \textbf{63.16} & \textbf{68.62} & \textbf{78.94} & \textbf{83.75} & \textbf{67.86} \\
\hline
\end{tabular}
\end{table}

\paragraph{Long-form generative tasks.}
Beyond short-answer MCQA, we evaluate \textbf{MAPLE} on GSM8K and BBH, which require multi-step chain-of-thought generation rather than single-token answer selection. Table~\ref{tab:generative} shows that at a 75\% budget on DeepSeek-MoE-16B, calibrated directly on each benchmark's chain-of-thought accuracy, \textbf{MAPLE} preserves GSM8K accuracy within 0.61 points of the full-budget baseline while reducing total evaluation time by 53.0\%, and improves BBH accuracy by 0.90 points while reducing total evaluation time by 6.3\%. These results indicate that \textbf{MAPLE}'s gains are not confined to short-form multiple-choice calibration, and that heterogeneous allocation remains competitive with, and can exceed, full-budget uniform routing on long-form generative reasoning.

\begin{table}[ht]
\caption{GSM8K and BBH accuracy and total evaluation wall-clock time at a 75\% budget, DeepSeek-MoE-16B, directly calibrated on each benchmark's chain-of-thought accuracy. GSM8K uses the complete 1{,}319-example test set (0-shot); BBH uses the complete 5{,}911-example set (3-shot CoT).}
\label{tab:generative}
\centering
\footnotesize
\setlength{\tabcolsep}{5pt}
\renewcommand{\arraystretch}{1.05}
\begin{tabular}{@{}lcccc@{}}
\hline
& \multicolumn{2}{c}{\textbf{GSM8K}} & \multicolumn{2}{c}{\textbf{BBH}} \\
\textbf{Configuration} & \textbf{Acc.} & \textbf{Total time} & \textbf{Acc.} & \textbf{Total time} \\
\hline
Full uniform (100\%) & 58.53 & 492.26\,s & 43.43 & 5{,}776.97\,s \\
MAPLE (75\%)          & 57.92 & \textbf{231.48\,s} & \textbf{44.33} & \textbf{5{,}434.15\,s} \\
$\Delta$              & $-0.61$ & $-53.0\%$ & $+0.90$ & $-6.3\%$ \\
\hline
\end{tabular}
\end{table}

\paragraph{Efficiency profile: offline cost versus serving gains.}
Table~\ref{tab:efficiency} reports both sides of \textbf{MAPLE}'s practical cost--benefit trade-off on DeepSeek-MoE-16B: panel~(a) gives the one-time offline calibration cost, incurred once per model before the resulting allocation is reused indefinitely at no further overhead; panels~(b) and (c) give measured serving latency and throughput in SGLang~0.4.6 and vLLM respectively, both on an ARC-Easy workload, where MoE layers assigned a single active expert use a dedicated execution path that skips multi-expert dispatch, and the model's dense layer is unaffected throughout. In vLLM, we additionally report a fused CUDA kernel variant of this $k=1$ path (``MAPLE 75\%''), which further reduces kernel-launch overhead. Gains under the two-GPU pipeline and expert parallelism are smaller than the single-GPU setting in SGLang, since communication and synchronisation overheads do not shrink in proportion to reduced expert computation.
\textbf{MAPLE} retains all expert weights throughout, so its benefit is reduced active computation and latency rather than model-size compression.

\begin{table}[ht]
\caption{\textbf{MAPLE}'s efficiency profile. (a) One-time offline calibration cost per model, summed across five tasks, on a single RTX A6000. (b) Measured SGLang~0.4.6 serving time on DeepSeek-MoE-16B at a 75\% budget, averaged over five runs, ARC-Easy workload. (c) Measured vLLM serving throughput and latency on DeepSeek-MoE-16B at a 75\% budget, ARC-Easy workload.
PP: pipeline parallelism; EP: expert parallelism.}
\label{tab:efficiency}
\centering
\begin{subtable}[t]{0.32\linewidth}
\centering
\scriptsize
\setlength{\tabcolsep}{2.5pt}
\renewcommand{\arraystretch}{1.05}
\begin{tabular}{@{}lc@{}}
\hline
\textbf{Model} & \textbf{Total (h)} \\
\hline
DeepSeek-MoE-16B  & 7.39  \\
DeepSeek-V2-Lite  & 14.86 \\
Moonlight-16B-A3B & 9.81  \\
OLMoE-1B-7B       & 6.10  \\
\hline
\end{tabular}
\caption{Offline cost}
\end{subtable}
\hfill
\begin{subtable}[t]{0.64\linewidth}
\centering
\scriptsize
\setlength{\tabcolsep}{3pt}
\renewcommand{\arraystretch}{1.05}
\begin{tabular}{@{}lccccc@{}}
\hline
\textbf{Setting} & \textbf{Full} & \textbf{Uniform, 75\%} & \textbf{MAPLE, 75\%} & \textbf{Time $\downarrow$} & \textbf{Tput. $\uparrow$} \\
\hline
1 GPU      & 2.860\,s & 2.680\,s & 1.940\,s & 32.2\% & 47.4\% \\
2 GPU, PP  & 3.660\,s & 3.660\,s & 3.470\,s & 5.2\%  & 5.5\%  \\
2 GPU, EP  & 9.056\,s & 8.809\,s & 8.701\,s & 3.9\%  & 4.1\%  \\
\hline
\end{tabular}
\caption{SGLang serving latency and throughput}
\end{subtable}

\vspace{4pt}
\begin{subtable}[t]{0.64\linewidth}
\centering
\scriptsize
\setlength{\tabcolsep}{3pt}
\renewcommand{\arraystretch}{1.05}
\begin{tabular}{@{}lccccc@{}}
\hline
\textbf{Setting} & \textbf{Full} & \textbf{Uniform, 75\%} & \textbf{MAPLE, 75\%} & \textbf{Time $\downarrow$} & \textbf{Tput. $\uparrow$} \\
\hline
1 GPU      & 3.915\,s & 3.745\,s & 3.736\,s & 4.58\% & 4.80\% \\
2 GPU, PP  & 13.703\,s & 11.819\,s & 11.762\,s & 14.16\% & 16.50\% \\
\hline
\end{tabular}
\caption{vLLM serving latency and throughput}
\end{subtable}
\end{table}

\subsection{Ablation Study}
\label{sec:ablation}

\paragraph{Sensitivity metrics.}
Table~\ref{tab:sensitivity_combo} compares different combinations of accuracy-range, weight-variance, and output-deviation sensitivity.

\begin{table}[h]
\caption{Ablation of sensitivity metrics (Acc.: accuracy range, Weight: weight variance, Output: output deviation). Best Gen. is the generation at which the best accuracy is reached.}
\label{tab:sensitivity_combo}
\centering
\scriptsize
\setlength{\tabcolsep}{3pt}
\renewcommand{\arraystretch}{1.05}
\begin{tabular}{@{}ccccccc@{}}
\hline
 & \textbf{Acc.} & \textbf{Weight} & \textbf{Output} & \textbf{Best Gen.} & \textbf{Best Acc.} & \textbf{$\Delta$} \\
\hline
1 &            &            &            & -- & 84.84 & --    \\
2 & \checkmark &            &            & 2  & 85.55 & +0.71 \\
3 & \checkmark & \checkmark &            & 4  & 85.20 & +0.36 \\
4 & \checkmark &            & \checkmark & 6  & 84.12 & -0.72 \\
5 &            & \checkmark & \checkmark & 10 & 83.75 & -1.08 \\
6 & \checkmark & \checkmark & \checkmark & 9  & 85.55 & +0.71 \\
\hline
\end{tabular}
\end{table}

\begin{figure}[htbp]
    \centering
    \includegraphics[width=\linewidth]{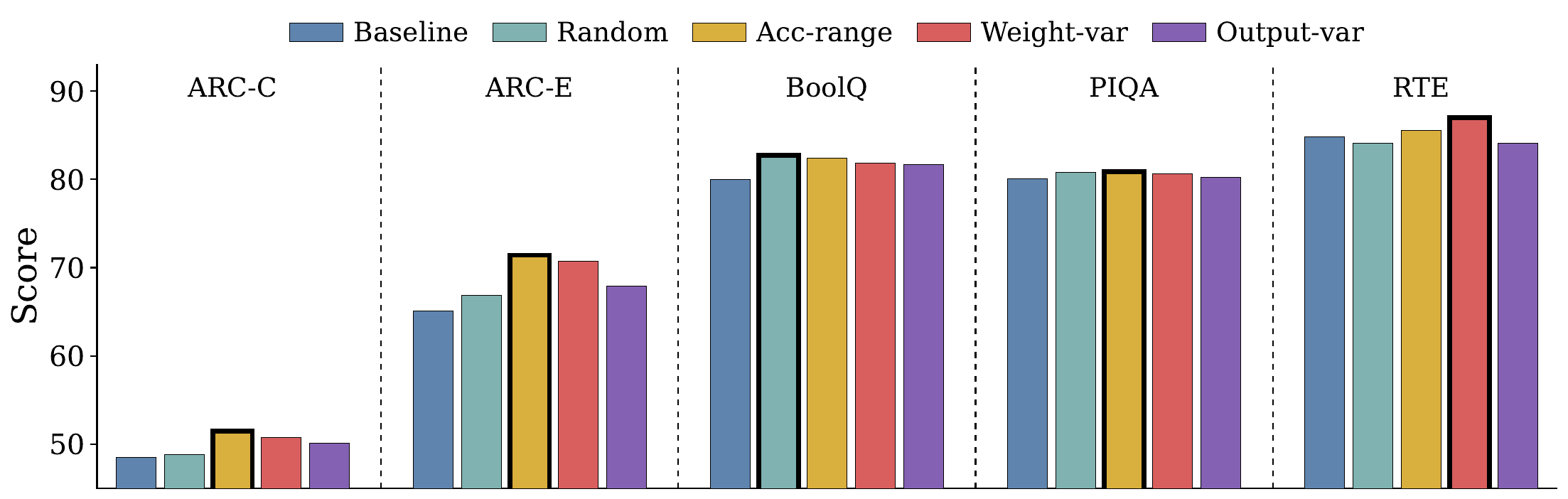}
    \caption{Performance of different sensitivity metrics in \textbf{MAPLE} under the 75\% expert budget across five tasks. Bold bar outlines indicate the best result within each task.}
    \label{dif_method}
\end{figure}

The full-uniform baseline reaches 84.84\% (row~1). Accuracy range alone (row~2) achieves the largest gain (+0.71) in just 2 generations. Adding weight variance (row~3) converges more slowly with a smaller gain (+0.36, gen.~4), while adding output deviation instead (row~4) hurts performance (-0.72, gen.~6). Weight and output variance without accuracy range (row~5) perform worst of all (-1.08, gen.~10), showing accuracy range does most of the useful work and the other two signals are redundant at best. Combining all three (row~6) matches the accuracy range alone (+0.71) but needs far more generations (9 vs.\ 2) to get there. We therefore use accuracy range as the default metric throughout the paper. Figure~\ref{dif_method} confirms this across tasks under the same 75\% budget, where accuracy range remains consistently competitive while weight and output variance are less stable. We therefore use accuracy range as the default metric.

\paragraph{Refinement and stability.}

Table~\ref{tab:first_batch_vs_overall} ablates the refinement step while keeping the budget and sensitivity metric fixed: refinement improves the closed-form allocation on every benchmark, with gains ranging from +0.82 on PIQA to +7.01 on ARC-E, showing that the closed-form solution provides an effective initial allocation while lightweight task-specific refinement yields further improvements. Since this stage is stochastic, we additionally rerun it with five independent seeds on DeepSeek-MoE-16B, fixing the model, calibration split, prompts, and expert budget, and varying only random initialisation;  the resulting seed standard deviation is an order of magnitude smaller than the refinement gain, indicating a stable solution rather than exploited noise.

\begin{figure}[htbp]
    \centering
    \begin{subfigure}[c]{0.495\linewidth}
        \centering
        \includegraphics[width=\linewidth,trim=8 8 8 8,clip]{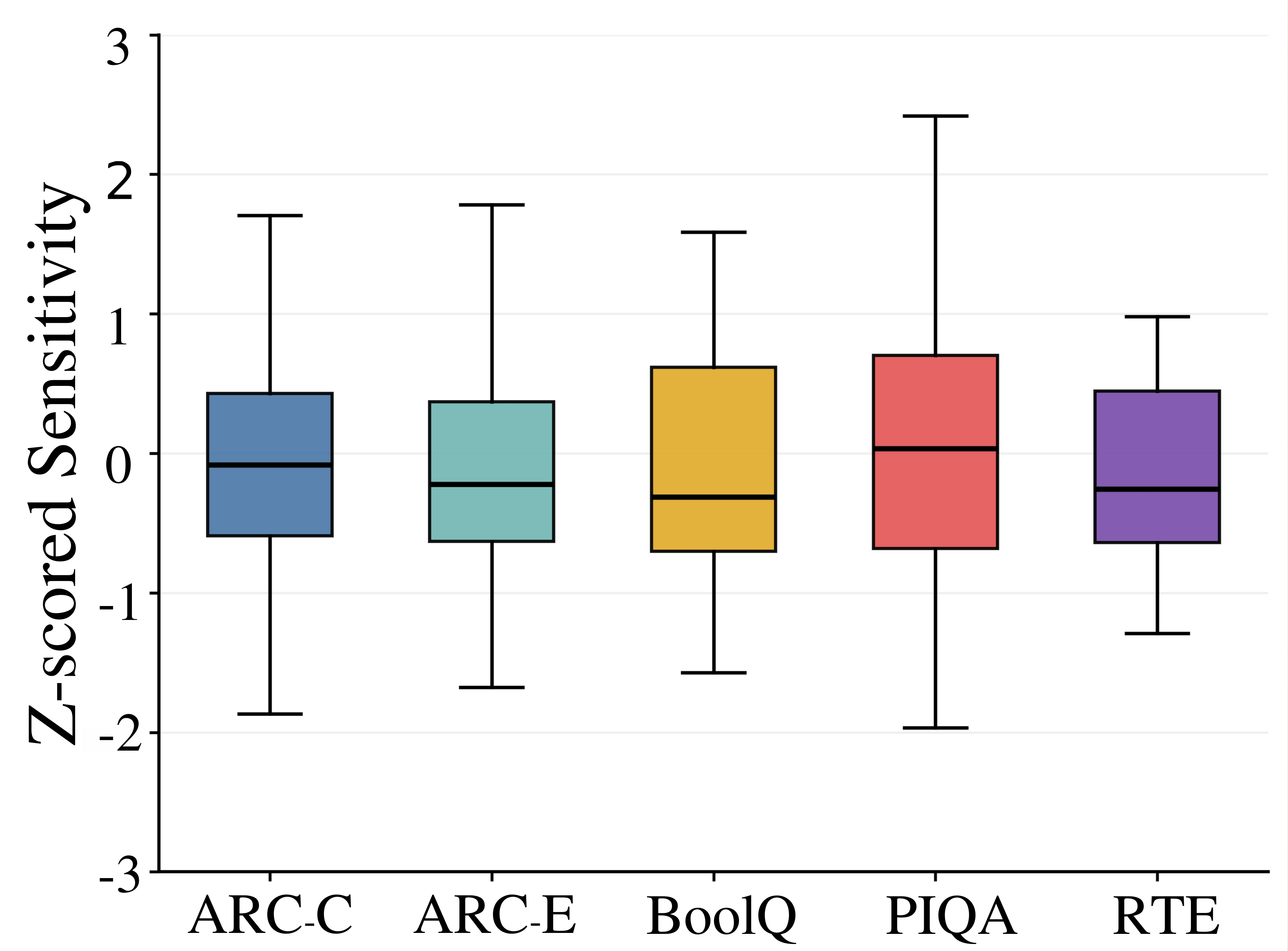}
        \caption{}
        \label{fig:a}
    \end{subfigure}
    \hfill
    \begin{subfigure}[c]{0.495\linewidth}
        \centering
        \includegraphics[width=\linewidth,trim=8 8 8 8,clip]{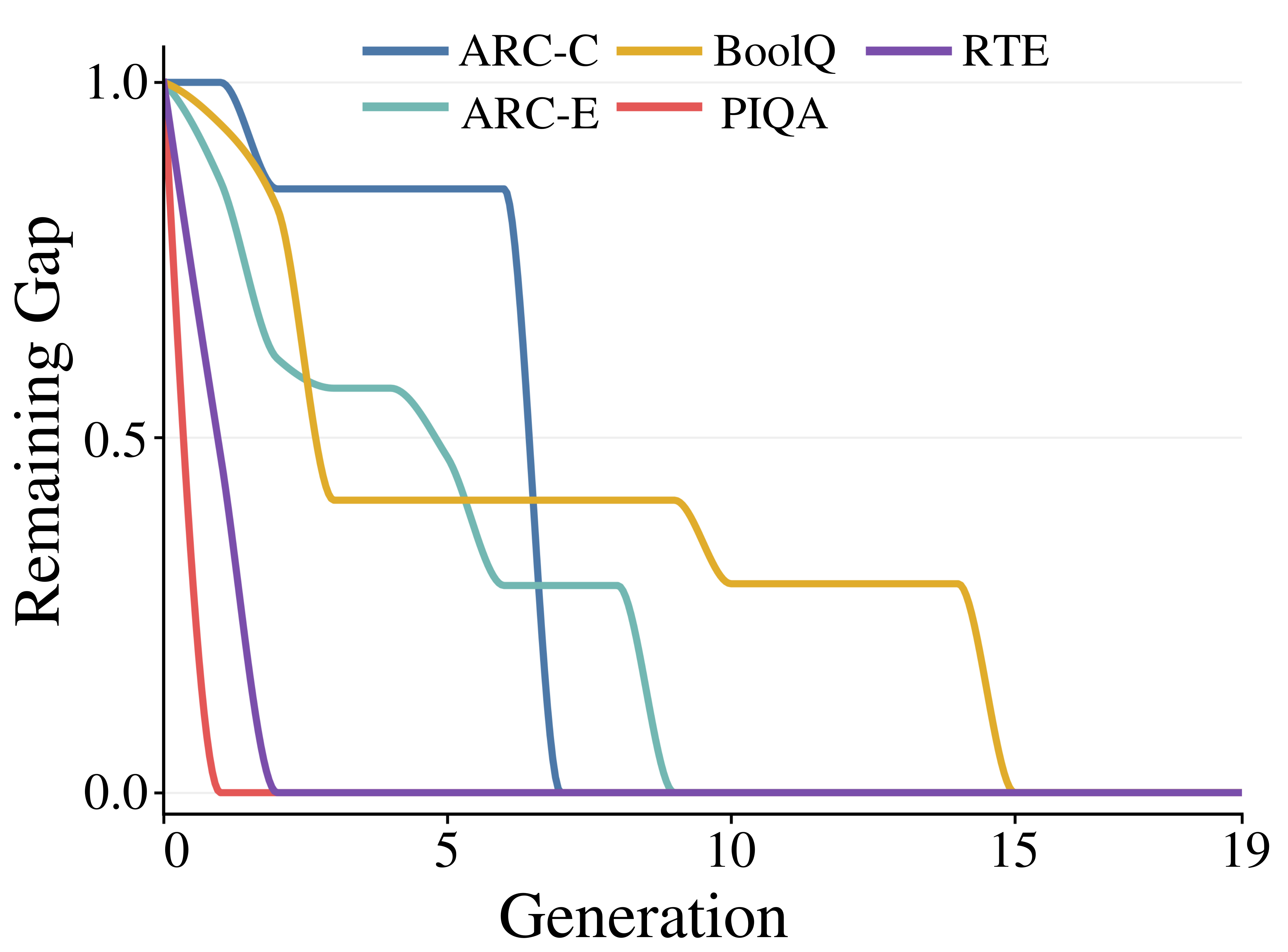}
        \caption{}
        \label{fig:b}
    \end{subfigure}
    \caption{(a) Task-wise standardised sensitivity distributions under the 75\% expert budget. (b) Normalised convergence trends across tasks, where Remaining Gap denotes the performance difference from the best-found solution.}
    \label{fig:combined}
\end{figure}

\begin{table}[htbp]
\caption{Effect of refinement and seed stability. Closed-form denotes the initial allocation, +Refinement applies the proposed refinement step, and Seed Std. reports the standard deviation of +Refinement accuracy across five independent refinement seeds.
}
\label{tab:first_batch_vs_overall}
\centering
\footnotesize
\setlength{\tabcolsep}{5pt}
\renewcommand{\arraystretch}{1.05}
\begin{tabular}{@{}lcccc@{}}
\hline
\textbf{Task} & \textbf{Closed-form} & \textbf{+Refinement} & \textbf{$\Delta$} & \textbf{Seed Std.} \\
\hline
ARC-C & 48.49 & 51.50 & +3.01 & $\pm 1.16$\\
ARC-E & 64.39 & 71.40 & +7.01 & $\pm 0.90$ \\
BoolQ & 80.70 & 82.39 & +1.68 & $\pm 0.45$\\
PIQA  & 80.03 & 80.85 & +0.82 & $\pm 0.21$\\
RTE   & 83.39 & 85.55 & +2.16 & $\pm 0.75$ \\
\hline
\end{tabular}
\end{table}
Figure~\ref{fig:combined} examines the refinement dynamics across tasks. Despite substantially different layer-wise sensitivity landscapes across benchmarks, most gains occur in the first few generations and the search converges quickly, supporting the use of a lightweight local refinement rather than an expensive global search.

\paragraph{Robustness of the calibration signal.}
Table~\ref{tab:calib_robustness} tests whether \textbf{MAPLE}'s gains depend on precise, task-specific calibration, or hold under weaker calibration signals on DeepSeek-MoE-16B at a 75\% budget.

A single allocation jointly calibrated across all five tasks (\emph{Reusable}) still delivers strong accuracy without any per-task recalibration, showing \textbf{MAPLE} does not require a fresh search for every deployment task.

\textbf{MAPLE} also transfers well across tasks: applying each task's searched allocation directly to the other four (\emph{Cross-task transfer}, averaged over the 20 off-diagonal source--target pairs shown in the lower rows) performs on par with established pruning methods despite never being calibrated on the target task, indicating that the sensitivity patterns \textbf{MAPLE} identifies reflect genuine, transferable properties of the model's layers rather than fitting narrowly to one evaluation set.

As an adversarial control, we replace task accuracy with negative token-level NLL on two generic corpora, C4 and FineWeb-Edu, sharing no examples, labels, or format with any evaluation task. Taken together, these results show \textbf{MAPLE} is robust along three independent axes: it does not need per-task recalibration, its allocations generalise across tasks, and its gains are not an artefact of the search procedure.

\begin{table}[ht]
\caption{Robustness of \textbf{MAPLE}'s calibration signal on DeepSeek-MoE-16B at a 75\% budget. Reusable, Cross-task transfer, C4-NLL, and FineWeb-Edu-NLL are alternative calibration signals; C4-NLL/FineWeb-Edu-NLL replace task accuracy with generic-corpus perplexity. The remaining rows give the full cross-task transfer matrix underlying the Cross-task transfer average above.
}
\label{tab:calib_robustness}
\centering
\footnotesize
\setlength{\tabcolsep}{4pt}
\renewcommand{\arraystretch}{1.05}
\begin{tabular}{@{}lcccccc@{}}
\hline
\textbf{Calibration signal} & \textbf{ARC-C} & \textbf{ARC-E} & \textbf{BoolQ} & \textbf{PIQA} & \textbf{RTE} & \textbf{Avg.} \\
\hline
Reusable (joint 5-task)                  & 49.83 & 68.95 & 82.05 & 78.62 & 83.39 & 72.57 \\
Cross-task transfer (avg.\ of 20 pairs) & 43.65 & 65.09 & 78.83 & 78.81 & 79.33 & 69.14 \\
C4-NLL                                   & 44.15 & 65.26 & 78.99 & 79.43 & 80.14 & 69.59 \\
FineWeb-Edu-NLL                          & 44.48 & 62.11 & 76.45 & 79.76 & 81.59 & 68.88 \\
ARC-C (source) & 51.50 & 64.56 & 79.24 & 78.40 & 79.06 & 70.55 \\
ARC-E (source) & 44.82 & 71.40 & 78.75 & 78.67 & 76.90 & 70.11 \\
BoolQ (source) & 40.80 & 64.56 & 82.38 & 79.16 & 81.95 & 69.77 \\
PIQA (source)  & 44.48 & 64.74 & 79.14 & 80.85 & 79.42 & 69.73 \\
RTE (source)   & 44.48 & 66.49 & 78.17 & 79.00 & 85.55 & 70.74 \\
\hline
\end{tabular}
\end{table}
\paragraph{Expert budget.}

\textbf{MAPLE}'s advantage over Wanda and SparseGPT holds at every budget level and widens at the tighter 50\% budget, suggesting heterogeneous reallocation becomes more valuable as naive pruning is more likely to remove capacity from layers that still need it. Heterogeneous allocation also improves over the uniform baseline at 100\% budget, indicating uniform allocation itself leaves room for better capacity placement independent of any budget reduction. 

\begin{table}[htbp]
\caption{ARC-C / RTE accuracy under 50\%, 75\%, and 100\% expert budgets on DeepSeek-MoE-16B. Wanda and SparseGPT are omitted at 100\% since their unpruned configuration is identical to the uniform baseline.}
\label{tab:reallocation_budget_compare1}
\centering
\footnotesize
\setlength{\tabcolsep}{8pt}
\renewcommand{\arraystretch}{1.05}
\begin{tabular}{@{}lccc@{}}
\hline
\textbf{Budget} & \textbf{Wanda} & \textbf{SparseGPT} & \textbf{MAPLE} \\
\hline
50\%  & 41.14 / 72.92 & 39.46 / 72.20 & \textbf{46.49 / 79.78} \\
75\%  & 45.81 / 82.67 & 46.15 / 81.22 & \textbf{51.50 / 85.55} \\
100\% (uniform baseline: 48.49 / 84.84) & -- & -- & \textbf{51.84 / 88.09} \\
\hline
\end{tabular}
\end{table}

\section{Conclusions and Limitations}
We presented \textbf{MAPLE}, a plug-and-play framework for heterogeneous expert allocation in budget-constrained MoE Transformers. \textbf{MAPLE} estimates layer-wise sensitivity, derives a closed-form allocation, and refines it with a lightweight genetic search, without weight updates, retraining, or architectural modification. Experiments on four MoE models and five benchmarks show that \textbf{MAPLE} outperforms pruning-based baselines under a 75\% routed-expert budget, and surpasses the full-budget uniform configuration on three models while using fewer routed experts. These results indicate that uniform expert allocation is often suboptimal, and that layer-wise capacity redistribution can better match the heterogeneous demands of MoE layers.

In the future, we plan to evaluate \textbf{MAPLE} on diverse hardware platforms, including GPUs, CPUs, and edge devices, and under broader real-world inference scenarios. This will allow us to further assess its practical efficiency, deployment flexibility, and applicability beyond the current experimental setting. In particular, hardware-aware evaluation can reveal whether reduced routed-expert computation translates into actual latency, memory, and energy benefits across different deployment environments. Such deployment-oriented analysis may also provide deeper insight into how heterogeneous expert allocation interacts with system-level constraints in practical inference systems, including memory bandwidth, batching behaviour, and hardware support for sparse MoE execution.

\bibliography{iclr2027_conference}
\bibliographystyle{iclr2027_conference}

\appendix
\section{Appendix}
You may include other additional sections here.

\subsection{Full Algorithm}
\label{app:algorithm}
Algorithm~\ref{alg:sensitivity_allocation} summarises the complete three-step \textbf{MAPLE} pipeline.
 
\begin{algorithm}[H]
\caption{Pseudocode for \textbf{MAPLE}}
\label{alg:sensitivity_allocation}
\KwIn{MoE model with $L$ layers; base expert count $k_0$; total budget $K$; valid range $[k_{\min}, k_{\max}]$; validation set $\mathcal{D}_{\mathrm{val}}$; population size $P$; generations $G$}
\KwOut{Final allocation $\mathbf{k}^{\mathrm{best}}$}

\BlankLine
\textbf{Step 1: Layer-wise Sensitivity Modelling}\;
\For{$l \leftarrow 1$ \KwTo $L$}{
    Probe layer $l$ by varying its expert count from $1$ to $k_0$, with all other layers fixed at $k_0$\;
    Record $A_l(e)$ for each candidate count $e$, and set the preferred count $k_l^* = \min\{e \mid A_l(e) = \max_{e'} A_l(e')\}$\;
    Compute the sensitivity score $s_l = \max_e A_l(e) - \min_e A_l(e)$ (accuracy range; see Section~\ref{sec:sensitivity} for alternative metrics)\;
}

\BlankLine
\textbf{Step 2: Sensitivity-guided Closed-form Expert Allocation}\;
\eIf{$\sum_{l=1}^{L} k_l^* < K$}{
    Let $\Delta = K - \sum_{l=1}^{L} k_l^*$\;
    Allocate the surplus to less sensitive layers: $k_l^{\mathrm{cont}} = k_l^* + \Delta \cdot \dfrac{1/s_l}{\sum_{j=1}^{L} 1/s_j}$ for each layer $l$\;
}{
    Let $\Gamma = \sum_{l=1}^{L} k_l^* - K$\;
    Remove the excess from less sensitive layers: $k_l^{\mathrm{cont}} = k_l^* - \Gamma \cdot \dfrac{1/s_l}{\sum_{j=1}^{L} 1/s_j}$ for each layer $l$\;
}
Clip $\mathbf{k}^{\mathrm{cont}}$ to $[k_{\min}, k_{\max}]$ and project it to integers via the largest-remainder method to obtain $\mathbf{k}^{\mathrm{sens}}$\;

\BlankLine
\textbf{Step 3: Sensitivity-constrained Genetic Refinement}\;
Initialise the population $\mathcal{P}$ around $\mathbf{k}^{\mathrm{sens}}$ by sampling $\tilde{k}_l = k_l^{\mathrm{sens}} + \epsilon_l$, with $\epsilon_l \sim \mathcal{N}(0,\sigma_l^2)$ and $\sigma_l^2 \propto \dfrac{1}{s_l+\varepsilon}$\;
\For{$g \leftarrow 1$ \KwTo $G$}{
    Evaluate each candidate $\mathbf{k} \in \mathcal{P}$ on $\mathcal{D}_{\mathrm{val}}$ and compute its fitness $f(\mathbf{k}) = A(\mathbf{k})$\;
    Keep the top-2 candidates in $\mathcal{P}$ as elites\;
    Refill $\mathcal{P}$ by perturbing elite solutions with sensitivity-controlled Gaussian noise, then project each result back to the feasible set\;
}
\Return{$\mathbf{k}^{\mathrm{best}} = \arg\max_{\mathbf{k} \in \mathcal{P}} f(\mathbf{k})$}\;
\end{algorithm}

\end{document}